\documentclass[draft=false,letterpaper]{article}
\usepackage[margin=0.75in]{geometry}
\usepackage[utf8]{inputenc}
\usepackage{algpseudocode}
\usepackage[round]{natbib}
\usepackage{hyperref}
\usepackage [english]{babel}
\usepackage [autostyle, english = american]{csquotes}
\usepackage{float}
\MakeOuterQuote{"}
\usepackage{graphicx}
\usepackage{float}

\usepackage[dvipsnames]{xcolor}
 
\usepackage{tikz}
\usetikzlibrary{positioning}

\title{Executive Function: A Contrastive Value Policy for Resampling and Relabeling Perceptions via Hindsight Summarization?
\\[0.25cm] \large Research Note}
\date{November 29, 2021}

\author{Chris Lengerich \\ \href{mailto:chris.lengerich@gmail.com}{chris.lengerich@gmail.com} \and Ben Lengerich \\ \href{mailto:blengeri@mit.edu}{blengeri@mit.edu}}

\begin{document}
\maketitle
\abstract{
We develop the few-shot continual learning task from first principles and hypothesize an evolutionary motivation and mechanism of action for executive function as a contrastive value policy which resamples and relabels perception data via hindsight summarization to minimize attended prediction error, similar to an online prompt engineering problem. This is made feasible  by the use of a memory policy and a pretrained network with inductive biases for a grammar of learning and is trained to maximize evolutionary survival. We show how this model of executive function can be used to implement hypothesis testing as a stream of consciousness and may explain observations of efficient human few-shot learning and neuroanatomy.
}

\tableofcontents

\section{Derivation of the Predictive Learning Task}

\footnotetext{The authors also thank MeiMei the golden retriever for insightful collaboration, in particular, deriving insights on human vs. dog data relabeling capabilities.}

\subsection{The Outer Loop: The Evolutionary Task}

Consider the online learning task of an agent in an environment such that the agent has a sensor system that produces a stream of time-ordered perceptions, $y_0, ..., y_t$. The agent lives completely within Plato's Cave \citep{plato_republic_375bc}, such that there are no labels for perceptions apart from implicit labels from future values of the stream, however, there is an evolutionary selection process based on an external environment which deletes agents based on their lack of fitness. The task we assign the agent is that of evolution - for the genetic code of the agent's species to continue to exist subject to this selection pressure.

\subsection{The Inner Loop: The Predictive Learning Task}
We consider in particular a \emph{predictive learning agent} - the agent which has a learned probability density function, $p(y_t|y_{t-1},...,{y_0})$, a prediction sampling process which samples $\hat{y_k}$ from $p(y_k|y_{t-1},...,{y_0})$ for a sampled $k >= t$, and an action space, $A$, from which our agent draws actions conditioned on its predictions. In environmentally fit agents, we expect the following desiderata for the agent's sampling policy and pdf:

\begin{enumerate}
    \item Predictive: The agent learns to calibrate its expectations accurately to future perceptions such that it minimizes the prediction loss, $\hat{y_k} - y_k$
    \item Controlled: The agent learn to predict calibrated ego-perceptions (perceptions which represent its own actions and thoughts) more often than other perceptions
    \item Aligned: The agent learns to predict calibrated ego-perceptions grounded to its survival more often
    \item Generalizable: The agent's predictions fulfill 1-3 across many environments and data distributions
\end{enumerate}

In contrast to the traditional TD Learning formulation of RL as a POMDP with a distinction between observations and rewards \citep{bellman_theory_1954, sutton_reinforcement_2018}, we treat reward simply as attention to a specific type of prediction loss between predictions and observations which is learned via evolutionary selection, in line with the view of dopamine's role as a generalized prediction loss signal within the brain rather than merely reward prediction \citep{gardner_rethinking_2018, hohwy_predictive_2014}.

\subsection{The Continual Few-Shot Learning Constraint}

We also introduce a few-shot learning constraint of only being able to collect a few examples of data from the environment. Continual few-shot learning allows for the agent's survival in environments with  distribution shift and expensive data collection, such as the competitive natural world. Although we will use tools of generational learning (as in most current machine learning setups now) to train agents, we are primarily interested in agents that display intrinsically motivated few-shot hypothesis testing behavior which can efficiently build a tree of knowledge.

Note that if we solve Desiderata 1 in generality, we have made substantial progress towards Desiderata 2 + 3, as these are just a conditional distribution over the joint distribution learned in 1 and can be fine-tuned with an exploration/exploitation policy and evolutionary selection mechanism.

\begin{figure*}[h]
    \centering
    \includegraphics[width=300px]{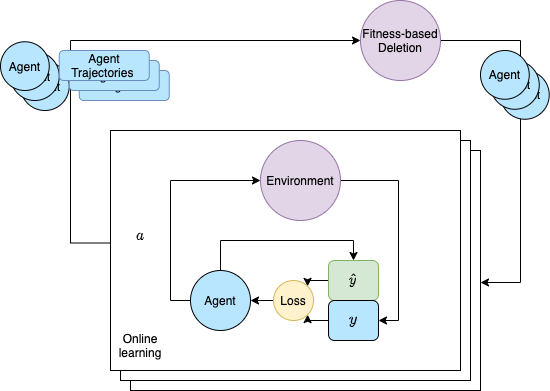}
    \caption{Learning Loops}
    \label{fig:two_loops}
\end{figure*}

\subsection{What does it mean to learn?}

Arguably, the goal of learning is to produce a \emph{world model} which is useful for prediction in environments with rapid change such as the competitive natural world \citep{ha_world_2018}. This requires the representations learned be \emph{sparse, compositional and causal}.\footnote{Note that although we would like world models to have accurate causal representations, even inaccurate causal models are useful for learning. Humans have been able to use approximate causal models of gravity effectively well before the discovery of Newton's law.}

\subsubsection{Sparsity}

Raw perception data is dense. However, in order to use it to predict, we would like to transform it into sparse\footnote{Sparsity is defined as density of zeros in a matrix} representations. Sparse representations are useful in that they can be converted into tokenized representations (vocabularies), which allow for compositionality and causal reasoning. 

\subsubsection{Compositionality}

We propose that it is particularly important to learn world models which have distributions of activations which are \emph{compositional in time} and \emph{compositional in abstractive depth}. 

\subsubsection{Compositionality in Time}
We define composition in time as mapping an event $A$ into an equivalent sequence of events $f(A) = a_0,a_1,...,a_n$ where $a_{i+1}$ occurs after $a_i$, as well as the inverse mapping  $g(a_0,a_1,...,a_n) = A$. This allows for the sequencing of events and long-horizon planning.

\subsubsection{Compositionality in Abstraction}
Composition in abstraction - that is, the ability to represent a concept either as an abstract idea (eg. "x = y") or a series of concrete examples (eg. "when x = 1, y = 1, when x = 2, y = 2, when x = 3, y = 3") - allows for the communication of concrete representations to other neural networks.  Mathematically, we would like to learn a mapping from an abstract concept $A$ to a series of concrete examples: $f(A) = a_0,a_1,...,a_n$, and an inverse mapping $g(a_0,a_1,...,a_n) = A$. Note that compositionality in abstraction is a softer variant of hierarchical representation as multiple mappings may exist.

\subsubsection{Communication is Composition in Time and Abstraction}

Composition may also be recognizable as communication. For example:

\begin{equation}
\verb!abstraction! \rightarrow \verb!1-3 concrete examples or stories! \rightarrow \verb!abstraction!
\end{equation}

This is the process of two trained neural networks (human brains) communicating abstractions between each other's networks using words. In the case of more shared parameters (common culture, familiarity), fewer words need to be said to transmit the same abstract idea between the two humans, and small changes in words can induce substantially different generalizations for humans \citep{luntz_words_2007}. As Chomsky has noted, the only indispensable operation of human grammar is that of Merge, which is not unlike compositionality in time and abstraction mentioned above \citep{chomsky_minimalist_1995}.

Communication, however, does not only occur externally to other humans using language. It also occurs to ourselves (via our inner voice) and between layers in neural networks using non-word representations. After all, outputs from a layer $L$ are latent variables which serve as inputs to layer $L+1$.  If you come from a machine learning background, you may notice that the diagram is simply the autoencoder setup shifted right by 1.

\subsubsection{Sparsity and Compositionality Are Necessary for Causality}

Sparsity, compositionality in time and compositionality in abstract representation are necessary properties of causal world models.

Although specifically how to learn causal world models is still an open research question, imparting inductive biases for sparse representations has been useful for improving neural network performance - key-value attention allows sparsity of communication in time \citep{vaswani_attention_2017}, while skip connections allow sparsity of communication in abstractive depth \citep{he_deep_2015}, since it allows lower layer world models to communicate directly with higher layers of a network.

\subsubsection{Contrastive Attention to Prediction Loss Drives Data Resampling and Relabeling Needed For Causal Models}

However, only having a sparse and compositional world model is not sufficient for it to be causal. For a world model to be causal it must be able to accurately model interventions and counterfactuals \citep{pearl_causality_2009}. For example, "does sunlight catalyze the production of ATP in plants?" requires understanding how the intervention (sunlight) affects your world model of plant energy generation.  Counterfactuals are a special type of intervention which is non-factual, for example, "without ATP synthase, would sunlight affect the production of ATP in plants?" If a model is able to answer a large number of these questions over a diversity of interesting interventions, we might consider that it has learned a causal model of the topic of sunlight and ATP.

\begin{figure*}[h!]
    \centering
    \includegraphics[width=200px]{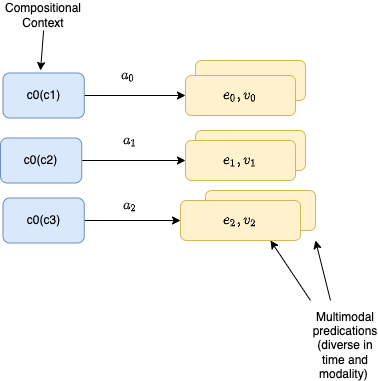}
    \caption{Compositional Contexts and Multi-Modal Predictions}
    \label{fig:hypothesis_testing}
\end{figure*}

Note that, with an existing world model of part of the input (say "sunlight"), it becomes easier to understand the question "does sunlight catalyze the production of ATP in plants?"  If you didn't know what sunlight means, on the other hand, your first instinct might be ask "what is sunlight?" to be able to systematically collect new training data about the part of the world model that was unknown. Similar contrastive prediction-loss motivated resampling has been shown to impart generalization with systematicity \citep{akyurek_learning_2021}, especially when inputs are collected for contrastive interventions and outputs have contrastive predictions induced by the same contextual data (see Figure \ref{fig:hypothesis_testing}).

Although there are many works describing useful inductive biases for neural network architectures trained on static training data, learning a policy for data labeling and data resampling over compositional representations needed for continual few-shot learning has not been well-addressed. To motivate this, we draw on observations of human learning, as humans are arguably experts in embodied continual few-shot learning that have behavior policies which have been pretrained by millenia of evolution.

\section{Biological Solutions}

Humans are efficient online learners who are often able to overcome the challenges of the continual learning environment and produce causal world models which are compositional in time and abstractive depth. Unlike many current machine learning models which run inference in serial, humans run learning, inference and control algorithms in parallel and may run training over the output perceptions from past training loops. Moreover, learning is tightly integrated with memory policies, unlike many contemporary RL agents.

\subsection{Prediction}

Humans are constantly making predictions about what perceptions they will experience in the future, $\hat{y}$. Predictions can be short-horizon (ie. where will my arm be in the next second) or long-horizon (what will I do in the next 5 years?). Predictions can be conscious or unconscious - at the conscious level, predictions are the goals, plans, tasks or imagination which drive behavior (see Figure \ref{fig:state}). At the unconscious level, they may appear as dreams \citep{freud_interpretation_1900}. 

\subsection{Prediction Loss}

When predictions are compared against later perceptions they create prediction loss, $loss(y,\hat{y})$. One concrete form of loss is just the L1 distance, $|\hat{y} - y|$.

In practice, we have more than one possible prediction loss in focus at a specific time, attention to which we mediate via executive function. Although executive function is often defined as the ability to set goals, plan and have impulse control, here we hypothesize a  simpler definition of executive function as stateful attention to a stack of predictions and perceptions, $loss(Y,\hat{Y}) = \sum a_i * loss(y_i, \hat{y_i})$ where $y_i$ is the $i$th perception on the stack, $\hat{y_i}$ is its prediction and $a_i$ is the  attention weight of the pair.  In classical RL terminology, this can be viewed as a long-horizon value policy for attention to predictive loss, which is then experienced as goals, plans and impulse control.

\subsubsection{Decreasing Prediction Loss}

Perceptions are subtly biased towards confirming predictions. Humans are prone to confirmation bias, whereby they create summaries which confirm our predictions without actually learning to control or predict accurately \citep{garrison_confirmation_1989}. At the subconscious level, we experience gestalt, subconsciously filling in missing parts of a whole \citep{wertheimer_untersuchungen_1923}.

However, when prediction loss is high enough, we experience this as an uncomfortable feeling of cognitive dissonance \citep{kaaronen_theory_2018}, and will take actions to reduce prediction loss by changing $y$ or $\hat{y}$. Anticipated prediction error can be reduced by taking actions within the external environment to change $y$ to match $\hat{y}$, while realized prediction error can be changed by modifying network weights so that $\hat{y}$ becomes closer to $y$ ("learning"), a behavior which is aided by using executive function to resample or relabel perceptions (changing $y$ and $\hat{y}$) ("thinking").

\subsubsection{Increasing Prediction Loss}

Although reducing prediction loss is a key drive, if we experience too little prediction loss, we also strive to increase it via curiosity-driven exploration, thus staying balanced within a "golden mean" of error \citep{plato_republic_375bc} formed by a Markov blanket over our predictions.\footnote{In regimes with high prediction loss, there is chaos, fear and uncomfortable uncertainty. In regimes with low prediction loss, boredom. In the middle is happiness. As in folk wisdom, "the journey is the reward."}

\subsubsection{Acting to Minimize Expected Prediction Loss}

\begin{figure*}[h]
    \centering
    \includegraphics[width=150px]{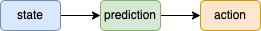}
    \caption{Active Inference}
    \label{fig:state}
\end{figure*}

When a prediction involves a ego-perception (ie. something that we will do in the future), in order to minimize expected predictive loss, we attempt to control our environment through action, a process called active inference  \citep{friston_free_2006} (Figure \ref{fig:state}). This can be consciously shaped by summarization and visualization - in sports, positive self-talk and visualization is a technique which is correlated with improved performance by creating prediction loss with respect to target performance \citep{raalte_self-talk_2017}.

Both conscious and unconscious predictions motivate action, regardless of their origin - ideas that "pop into your head" tend to have connections to past events. For example, when you have just read the sentence "does sunlight catalyze the production of ATP in plants?", you may be more likely to eat a salad tonight or go outside \citep{luntz_words_2007}.

\subsection{Learning to Minimize Prediction Loss}

\subsubsection{What is Learning?}

The most straightforward way to minimize realized  prediction loss is by updating weights or growing neural connections such that they directly change the world model so that the predictions are closer to perceptions. In artificial neural networks this occurs via backpropagation \citep{rumelhart_learning_1986} or many other optimization methods, and in the human brain, via Hebbian learning \citep{hebb_organization_1949}.

\subsubsection{Taking actions in the world to resample data}

To assist with learning, as with most animals, we can resample data via physical movement which results in new contrastive perceptions \citep{friston_predictive_2009}. For example, at a cocktail party, you may move your head to look at a speaker's lips to be able to better distinguish the utterances of the speaker from the background noise. We can also employ longer-term learning loops to resample data, for example, calling a friend for a second opinion on how to approach a situation. In this case, you acquire two summary labels for the situation (one from yourself, one from your friend). This contrastive data makes it easier to learn accurate causal world models.

\subsubsection{Taking internal actions to resample data}

Although taking actions is an effective way to gather contrastive data, it is also quite expensive and impractical in many settings. Shifting attention to novel internal perceptions allows us to sample new perceptions and predictions without motor control. Humans have learned to hack this process for pleasure and control. Attending to the sensations of your body which are easily predictable via meditation may be a method to consciously lower prediction loss by changing the contents of the prediction stack  \citep{jamieson_unified_2015, pagnoni_contemplative_2019}. Mindfulness lowers reward prediction error and activates the putamen in practicioners \citep{kirk_mindfulness_2015}. Consciousness presence has been hypothesized to be due to top-down predictions successfully suppressing informative interoceptive signals \citep{seth_interoceptive_2012}.

\subsection{Few-Shot Continual Learning}

However, even with policies for data resampling, learning a causal world model from a handful of examples and random neural network weights tends to be infeasible. Humans, however, benefit from evolutionarily trained data encoded in the inductive biases of neuroanatomy. Learning policies which tap into these existing facilities are much faster than learning without. 

\subsubsection{A Grammar of Learning?}

The Complementary Learning Systems theory \citep{mcclelland_why_1995} proposes that the hippocampus encodes "fast" learning policies resulting in direct storage of incoming perceptions, while the neocortex replays these perceptions slowly (interspersed with other perceptions) to achieve a slow learning of abstractions and subconscious behaviors.

We propose that fast learning may be a behavior that uses the tools of summarization, active inference and multiple stateful passes through a pre-trained network (which activate latent abstractions) such that the resulting output data forms a batch which has sparse and compositional representations such that it can be more easily learned, a phenomenon we call learning a communicative mapping to a \emph{grammar of learning}, similar to an online prompt engineering problem. Specifically, for any network $\theta$ which has been pre-trained on dataset $D$, there exists a data distribution $D'$ such that an input $d' \in D'$ will produce a sparse, compositional distribution of outputs which can be resampled and relabeled contrastively, thus inducing efficient learning. We call this $D'$ a grammar of learning for the network and the original grammar of learning without pre-training data a universal grammar of learning \footnote{This is largely just an inductive bias for sparse and compositional input data}. Given a novel input $e$, if the network can find a series of transformations to map $e$ into a latent representation $d'$, then $e$ can be more easily learned by association with $d'$.

\subsubsection{Resampling and Relabeling Data Via Memory and Hindsight Summarization}

\begin{figure*}[h!]
    \centering
    \includegraphics[width=400px]{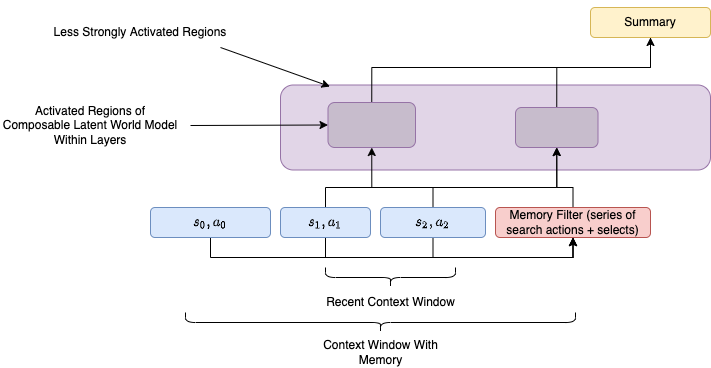}
    \caption{Relabeling Data Via Hindsight Summarization}
    \label{fig:hindsight_summarization}
\end{figure*}

\begin{figure*}[h!]
    \centering
    \includegraphics[width=400px]{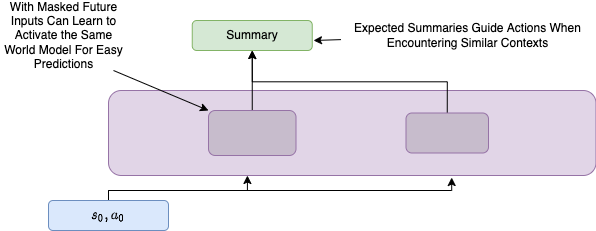}
    \caption{Continuous Training on Relabeled Data to Motivate Behavior To Connect Summaries to Future Actions}
    \label{fig:training_summaries}
\end{figure*}

The average human reaction time is ~250ms and the human eye has a resolution of ~576 megapixels, resulting in a data stream of at least ~7GB/second. To have the  compositional and sparse representations required for learning of causality, we must significantly reduce the dimensionality of this firehose of data.  This done by hindsight summarization (Figure \ref{fig:hindsight_summarization}). Hindsight summarization has access to a memory and a grammar of learning to enable rapid non-linear data transformations via existing world models. Creating a hindsight summary changes the composition of the perception stack attended by executive function, creating new prediction losses which can be used directly as training data (Figure \ref{fig:training_summaries}) or indirectly to produce further summaries, forming a stream of consciousness which quickly changes the composition of the entire perception stack.


 \clearpage

Streams of consciousness like beneficial self-talk \citep{raalte_self-talk_2017} and cognitive behavioral therapy \citep{manjaly_computational_2020} may be learned behaviors to modify experienced prediction loss using hindsight summarization.

These can be complex and diverse. Adopting a growth mindset by framing unpredictable perceptions as a hypothesis test (so that any outcome is now expected with a certain probability) is correlated with reduced cognitive dissonance during experimentation for the authors. Making explicit predictions before reading (a form of data relabeling via increasing attention to prediction loss) has been shown to improve human learning \citep{thomas-fair_power_2005, brod_predicting_2021}, and may be an example of consciously increasing prediction loss to focus learning mechanisms. They may also occur unconsciously - humans make long-range predictions at multiple levels of latent depth when reading \citep{caucheteux_long-range_2021}.\footnote{Eric Hayot, a comparative literature professor, once told one of the  authors to "kick the ladder" - that is, after finishing a draft, rewrite it with the benefit of hindsight so that you weave anticipation of the summary into the piece, and remove the confusing scratch writings you used to bootstrap the summary for yourself as an author. In this way, the reader's executive function  naturally anticipates the conclusion with less cognitive dissonance.}

Regions of the brain associated with executive function are larger in humans than other primates \citep{donahue_quantitative_2018}, and humans tend to be better at few-shot learning than other animals, which may be related to the ability to set conscious plans \citep{fuster_chapter_2017}.

Analogues to simple patterns of hindsight summarization for non-embodied agents can also be seen in SOTA approaches to self-supervised learning, such as data2vec \citep{baevski_data2vec_2022}, where summaries can be considered to be the target latent representations of the teacher, and the future masking of Figure \ref{fig:training_summaries} is analogous to the input masking in the student model in \cite{baevski_data2vec_2022}. However, data2vec does not use multiple steps of conditional computation, as humans do.

Hindsight summarization can also be compared to other hindsight schemes such as HER \citep{andrychowicz_hindsight_2018}, however  summarization is a learned path function over the past trajectories rather than a deterministic function of the last state, as in HER. Unlike generalized hindsight \citep{li_generalized_2020}, hindsight summarization encodes information about a trajectory of observations, not just rewards, and runs on a much faster timescale, possibly summarizing a sequence of perceptions which span intervening training epochs.

\subsubsection{Hypothesis testing via summarization}

\begin{figure*}[h!]
    \centering
    \includegraphics[width=450px]{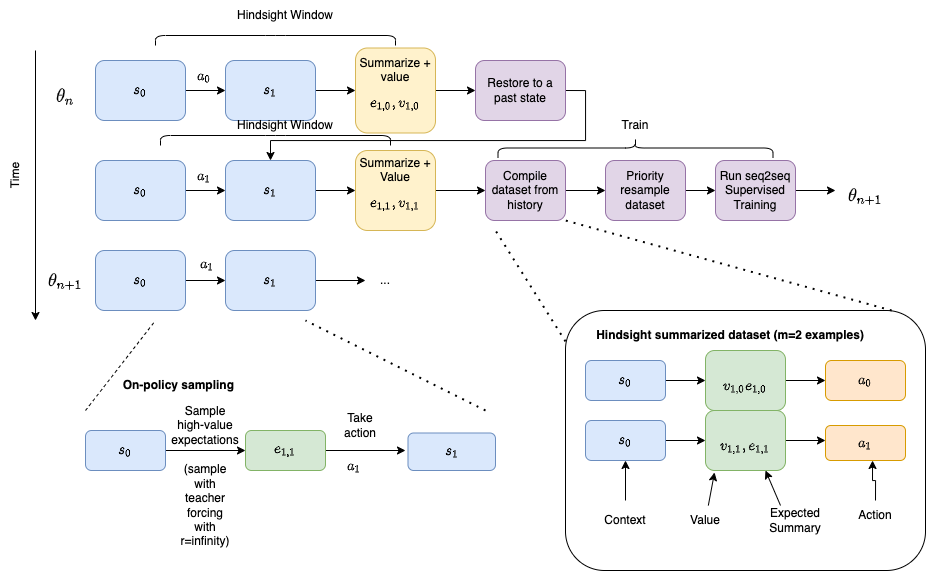}
    \caption{Hypothesis Testing}
    \label{fig:data_resampling}
\end{figure*}

Sequential steps of hindsight summarization with intervening training and counterfactual sampling may be recognizable as the Scientific Method, a well-known meta-learning algorithm for scientific discovery. See Figure \ref{fig:data_resampling} for how we might model this stream of consciousness and use insights to inform future behavior.

 \clearpage
 
\section{Conclusion}

We have defined the few-shot continual learning task from first principles and hypothesized an evolutionary motivation and mechanism of action for executive function as a contrastive value policy which resamples and relabels perception data via hindsight summarization to minimize attended prediction error, similar to an online prompt engineering problem. This is enabled by the use of a memory policy and a pretrained network with inductive biases for a grammar of learning and is trained to maximize long-term survival. We have shown how this model of executive function can be used to implement hypothesis testing and may explain observations of few-shot learning and neuroanatomy.

There is ample room for future work. Hypothesis testing is only one example of a stream of consciousness that can be implemented using this architecture. In contrast, humans are experts at using complicated patterns of hindsight summarization and intervening training and action for various meta-learning strategies. As a simple example, how to implement imitation learning where a student learns not only to copy a teacher, but who to trust within a competitive environment?

The parallels between human learning and machine learning also remain to be explored more rigorously. Although we have shown correlations in empirical observations and a hypothesized mechanism of action, to show causality and biologically relevant mechanisms requires additional study. We anticipate isomorphism in the objective functions, training data and behavior of in silico and in vitro algorithms but not necessarily the specific architectures. For example, the brain implements information processing algorithms in spiking neurons and has evolved distinctive regions that may not necessarily be seen in machine learning models. We do, however, see concrete opportunities to use isomorphism in purpose and algorithmic structure to better inform our understanding of learning in both humans and machines.

\footnotetext{The authors thank Volodymyr Kuleshov, Pang Wei Koh, Lili Chen, Michael Janner, Andrej Karpathy, Laura Deming, Minn Kim, Maarten Bosma, Ruoxi Wang, Chip Huyen, Kanjun Qiu, Eugene Lengerich, Rebecca Lengerich and Charles Antle for insightful feedback on drafts of this work. Empirical experiments were run on OpenAI's fine-tuning API.}

\bibliographystyle{plainnat} % We choose the "plain" reference style
\bibliography{refs}

\end{document}